\documentclass{article}

\usepackage[preprint]{neurips_2026}
\usepackage{amsmath}
\usepackage{multirow}
\usepackage{graphicx}
\usepackage{titletoc}
\usepackage{booktabs}
\usepackage{tabularx}
\usepackage{array}
\usepackage{hyperref}
\usepackage{amsmath}
\usepackage{amssymb}
\usepackage{algorithm}
\usepackage{algorithmicx}
\usepackage{algpseudocode}

\usepackage[utf8]{inputenc} % allow utf-8 input
\usepackage[T1]{fontenc}    % use 8-bit T1 fonts
\usepackage{hyperref}       % hyperlinks
\usepackage{url}            % simple URL typesetting
\usepackage{booktabs}       % professional-quality tables
\usepackage{amsfonts}       % blackboard math symbols
\usepackage{nicefrac}       % compact symbols for 1/2, etc.
\usepackage{microtype}      % microtypography
\usepackage{xcolor}         % colors
\usepackage{enumitem}
\usepackage{booktabs}
\usepackage{caption}
\usepackage{graphicx}
\usepackage{wrapfig}
\title{No Magic, Just Building Blocks: Sparse Compositional Flow Matching for Embodied Trajectories}

\author{%
  Yan Tang\textsuperscript{1}, Yuanbo Tang\textsuperscript{1}, Tingyu Cao\textsuperscript{1}, Shaolun Huang\textsuperscript{1}, 
  Yang Li\textsuperscript{2} \\
  \textsuperscript{1} Tsinghua Shenzhen Graduate School, Tsinghua University,  Shenzhen, China \\
  \texttt{tangyan7025@gmail.com} \\
  \textsuperscript{2} School of AI, Chinese University of Hong Kong (Shenzhen) \\
  \texttt{yangl@cuhk.edu.cn}
}

\begin{document}

\maketitle
%标题更专一些，指具身领域和轨迹领域
%摘要太术语化了，变得更泛一些
\begin{abstract}
Embodied trajectories, such as the executable motion sequences of robotic manipulators, underwater vehicles, and mobile robots, are a fundamental output of embodied AI. Modern generative models often treat them as a dense, monolithic signal generated point by point, fitting an intricate high-dimensional posterior while leaving the data's latent structure unmodeled, the same sample inefficiency long identified by the structured generative model literature. We argue that a \emph{compositional latent structure} is a natural choice: many embodied tasks share recurring motion fragments that can be made explicit as a finite repertoire of reusable motion primitives, and compositional units naturally align with subtask boundaries to support task decomposition. Existing compositional generators, however, compose in a latent space and rely on post-hoc decoding to relate sampled units to actual trajectory segments. We instead compose \emph{directly in the physical trajectory space} through a flow-matching framework with two coupled designs. \emph{Motion-Primitive Dictionary Learning} equips each atom with a learnable length mask and binary starting indicators so the atom itself is the primitive, reused verbatim wherever it is placed. \emph{Structural Sparse Flow Matching with Geometric Constraints} then generates a binary placement matrix using duration-aware tokenization and a differentiable geometric loss that enforces spatial continuity and temporal contiguity where adjacent primitives meet. On Open X-Embodiment and 3DMoTraj, the framework attains state-of-the-art accuracy and reduces the FDE/ADE ratio from ${\sim}1.8$ to $1.07$, improving ADE by 19.2\% and FDE by 21.0\% over the strongest baseline.% Anonymous code is available at \url{https://anonymous.4open.science/r/SSGFLOW-621A/}
\end{abstract}
\section{Introduction}

Embodied trajectories are the executable interface of physical agents: robotic manipulators, autonomous underwater vehicles navigating turbulent flows, and mobile robots traversing cluttered environments all ultimately commit to a sequence of physically realizable motions~\cite{10912754,10.5555/3780338.3783536,yu2024trajectory}. Generating such trajectories is therefore a fundamental task that underpins planning~\cite{ajay2023compositional}, control~\cite{10.1177/02783649241273668}, and skill transfer~\cite{NEURIPS2023_c276c330} in embodied AI.

Modern generative models like VAEs, diffusion and flow matching~\cite{Xiang_YIN_Wang_Jin_2024,NEURIPS2023_cd9b4a28,11095073,hou2026cfo} have substantially advanced trajectory modeling, yet often share a single sample-space convention: a trajectory is treated as a \emph{dense, monolithic} signal generated point by point in continuous coordinate space. Generating trajectory points in this way faces the same fundamental difficulty as generating any other complex high-dimensional data: the joint posterior is intricate enough that estimating it accurately demands large amounts of training data, while latent structure within the data is left unmodeled. This is precisely the gap addressed by the \emph{structured generative model} literature~\cite{lin2026partcrafter,favero2025how}, which advocates explicitly modeling latent structure to reduce the sample complexity of generation. The remaining open question is therefore not whether trajectory generation needs a latent structure, but \textbf{what form of latent structure is appropriate}.

Among possible latent structures, \emph{compositional structure} is a natural choice for embodied trajectories. \textbf{(i)~Task commonality}: tasks such as reaching, grasping, releasing, and retracting share recurring motion fragments within and across tasks, supporting an explicit, finite repertoire of reusable motion primitives. \textbf{(ii)~Subtask alignment for interpretability}: compositional units naturally align with subtask boundaries, making the representation directly usable for task decomposition without post-hoc clustering or attribution. Existing compositional generators for embodied behavior, however, compose in a \emph{latent space}, over skills, options, latent actions, or task skeletons~\cite{ajay2023compositional,Wang-RSS-24,NEURIPS2023_c276c330,NEURIPS2024_82389fbf,NEURIPS2024_e51ec472}, and rely on post-hoc decoding or attribution to relate sampled units to actual trajectory segments, so the property that a sampled element \emph{directly is} an identifiable trajectory segment remains an aspiration. 

\begin{figure}[t]
  \centering
  \includegraphics[width=0.95\linewidth]{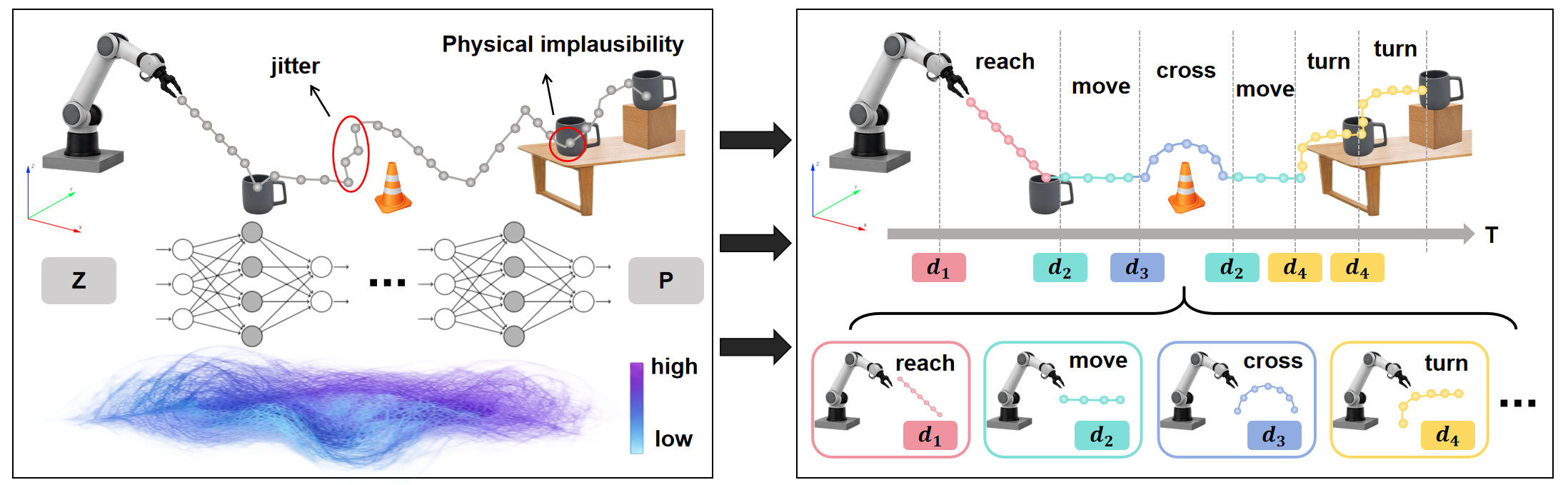}
  \caption{Compositional structure as the latent structure of embodied trajectories. Compared to dense, pointwise generation (left), our framework (right) reformulates trajectory generation as the placement of a finite library of reusable motion primitives on a shared timeline, exposing the latent structure of embodied trajectories as identifiable, reusable units with intrinsic interpretability.}
  \label{fig:fig1}
\end{figure}

\textbf{Composing in physical trajectory space, not latent space.} We instantiate this idea by maintaining a dictionary $\tilde{\mathbf{D}}$ of motion-primitive segments and assembling a full trajectory by \emph{geometrically tiling} selected primitives onto a shared timeline. This is non-trivial: each entry must remain shape-stable across instances (standard convolutional dictionary learning fails here, as atom shape and placement are jointly absorbed into the convolution output), and adjacent primitives must meet at their boundaries in a geometrically valid way. We resolve both with two designs sharing a single placement variable. \textbf{Motion-Primitive Dictionary Learning~(MPDL)} equips each atom with a learnable length mask and binary starting indicators, so the atom \emph{itself} (not its convolution output) is the primitive and appears verbatim wherever placed. \textbf{Structural Sparse Flow Matching with Geometric Constraints} then generates the placement matrix $\hat{\mathbf{R}}\!\in\!\{0,1\}^{M\times L}$, which primitives start at which timesteps, using duration-aware tokenization and a differentiable geometric loss on the predicted clean endpoint that penalizes spatial discontinuities and temporal gaps at primitive boundaries. On Open X-Embodiment\cite{open_x_embodiment_rt_x_2023} (manipulation) and 3DMoTraj~\cite{10.5555/3780338.3783536} (3D underwater) benchmarks, the framework attains state-of-the-art accuracy and reduces the ratio of final displacement error (FDE) to average displacement error (ADE) from ${\sim}1.8$ to $1.07$, improving ADE by 19.2\% and FDE by 21.0\% over the strongest baseline.

\noindent\textbf{Contributions.}
\begin{enumerate}[leftmargin=*,itemsep=1pt,topsep=2pt]
    \item A trajectory generation framework that composes \emph{in the physical trajectory space} rather than a latent space, so every generated unit is itself an identifiable trajectory segment without post-hoc decoding or attribution.
    \item \emph{Motion-Primitive Dictionary Learning}: anchored differentiable length masks and binary starting indicators yield a primitive library whose entries remain shape-stable across instances and are directly samplable by a generative model.
    \item \emph{Structural Sparse Flow Matching with Geometric Constraints}: flow matching over the binary placement matrix with duration-aware tokenization and a differentiable geometric loss on the predicted clean endpoint, achieving an FDE/ADE ratio of $1.07$ (vs.\ ${\sim}1.8$ for baselines) and 19.2\%\,/\,21.0\% ADE\,/\,FDE gains on two embodied benchmarks.
\end{enumerate}
\section{Related Work}

\paragraph{Generative trajectory models.}
Trajectory generation has been studied across pedestrian forecasting~\cite{10656710,10378344,10547191}, autonomous driving~\cite{11092858,11444908,10.1016/j.eswa.2025.128545,10271561}, robot manipulation~\cite{10912754,10.1177/02783649241273668,10611507}, and 3D embodied motion~\cite{10.5555/3780338.3783536,10.5555/3666122.3667002}, with modeling families spanning RNN/Transformer regressors~\cite{10678069,10378344,8954402}, CVAEs~\cite{Xiang_YIN_Wang_Jin_2024,9577622,9286482}, GANs~\cite{10377513,10.1016/j.future.2024.07.011,10592580}, normalizing flows~\cite{NEURIPS2024_69f3eb24,9515316}, denoising diffusion~\cite{NEURIPS2023_cd9b4a28,luo2025generative,10203059}, and flow matching~\cite{11095073,10.5555/3737916.3741320,kornilov2024optimal,hou2026cfo}. Despite this methodological diversity, these models share a single sample space, a dense coordinate sequence whose elementary unit is a single timestep, leaving motion structure and segment boundaries implicit. We do not propose another dense estimator; instead, we change the sample space itself, generating trajectories as sparse placement over a learned primitive library.

\paragraph{Motion primitives and dictionary learning.}
A long line of work represents motion as compositions of reusable units, including Dynamic and Probabilistic Movement Primitives~\cite{10050558,NIPS2013_e53a0a29,10.5555/3495724.3496149}, options~\cite{NEURIPS2024_e51ec472}, skill embeddings~\cite{NEURIPS2023_c276c330}, latent-action models~\cite{NEURIPS2024_82389fbf}, and motif-based representations~\cite{10.1145/3272127.3275038}. In parallel, sparse coding and convolutional dictionary learning~\cite{9577340} represent temporal signals as shifted atoms with sparse activations, with applications to motion segmentation and trajectory compression~\cite{10.1145/3589132.3625607}. Across both threads, the learned units have been used predominantly for representation, retrieval, or control rather than as the sample space of a generative process. We close this gap by attaching a differentiable length boundary and a multi-hot binary onset to each atom, so that the atom \emph{itself} is the primitive and is directly samplable by a generative model.

\paragraph{Compositional and structured generative models.}
Compositional generative models have been studied for visual synthesis~\cite{10.1007/978-3-031-19790-1_26,11093619}, human motion~\cite{11093839}, traffic simulation~\cite{11092842}, robotic planning~\cite{ajay2023compositional}, and policy learning~\cite{Wang-RSS-24}, with diffusion- and flow-based variants supporting guided sampling, condition conjunction, and trajectory stitching across complex distributions~\cite{10.5555/3618408.3618749,NEURIPS2025_52d36bc3,luo2025generative}. More broadly, the \emph{structured generative model} literature~\cite{lin2026partcrafter,favero2025how} advocates explicitly modeling latent structure to reduce the sample complexity of generation. In nearly all of this work, however, composition operates on conditions, latent factors, or predefined task skeletons, while the generated object itself remains a dense sequence whose elementary units are not directly exposed. Our framework departs structurally: composition \emph{is} the sample space; the model generates discrete primitive--time placements and regularizes the predicted clean endpoint with a differentiable geometric loss that enforces spatial continuity and temporal contiguity between adjacent units.

\section{Method}
\label{sec:method}

We model embodied trajectory generation as sparse composition over a learned motion-primitive library, where every trajectory is explained by a sparse binary placement matrix $\mathbf{R}\!\in\!\{0,1\}^{M\times L}$ over a shared timeline.
Throughout this paper, $\mathbf{R}$ is the sole intermediate representation: the dictionary parameterizes how $\mathbf{R}$ reconstructs back to a trajectory, while the flow matching vector field parameterizes how noise transports onto the manifold of physically valid $\mathbf{R}$.
Both parameterizations are coupled through this shared variable and trained in a single optimization loop under one placement legality energy, so that dictionary learning and generative modeling never become separable modules.
Sec.~\ref{sec:dict} defines the primitives and compositional synthesis; Sec.~\ref{sec:joint} formalizes the joint loop; Sec.~\ref{sec:cond} instantiates it for conditional prediction.

%% ====================================================================
\subsection{Motion-Primitive Dictionary Learning}
\label{sec:dict}
\label{sec:cdl}
\label{sec:masked_cdl}

\paragraph{Notation.}
Let $\mathbf{X}\!=\!\{\mathbf{x}_i\}_{i=1}^{N}$ with $\mathbf{x}_i\!\in\!\mathbb{R}^{C\times L}$ denote $N$ trajectories of $L$ timesteps in $C$-dimensional Cartesian space, and let $\mathbf{D}\!=\!\{\mathbf{d}_j\}_{j=1}^{M}$ with $\mathbf{d}_j\!\in\!\mathbb{R}^{C\times K}$ be a convolutional dictionary of $M$ atoms of maximum temporal extent $K\!\le\!L$.
A standard convolutional dictionary paired with continuous-valued activations entangles atom shape with instance-specific placement, producing primitives whose realized form drifts across occurrences (full diagnosis in App.~\ref{app:pcdl_details}).
We therefore define primitives so that shape, duration, and placement are separated by construction.

\paragraph{Anchored masked atoms.}
Each atom $\mathbf{d}_j$ carries a temporal extent through an unconstrained parameter $\phi_j\!\in\!\mathbb{R}$.
We define a soft width and a sharpened-sigmoid length mask of steepness $\alpha$,
\begin{equation}\label{eq:soft_width}
  w_j \;=\; 1 + (K-1)\,\sigma(\phi_j)\;\in\;(1,\,K),
  \qquad
  m_{j,s} \;=\; \sigma\!\bigl(\alpha\,(w_j - s - \tfrac{1}{2})\bigr),\quad s = 0,\dots,K{-}1,
\end{equation}
and form the \emph{effective atom} $\tilde{\mathbf{d}}_j = \mathbf{d}_j\odot(\mathbf{1}_C\,\mathbf{m}_j^{\!\top})\in\mathbb{R}^{C\times K}$, a temporally contiguous trajectory segment of learnable duration anchored at $s{=}0$ in the atom's local frame.
A straight-through discretization yields integer effective lengths $\bar{w}_j\in\{1,\dots,K\}$ for downstream index arithmetic; the structural reason why this anchoring + length-mask combination is necessary is given in App.~\ref{app:why_anchor}.

\paragraph{Multi-hot binary placement.}
For trajectory $i$ and atom $j$, a logit vector $\boldsymbol{\ell}_{i,j}\!\in\!\mathbb{R}^{L}$ parameterizes independent onset probabilities $\mathbf{q}_{i,j}=\sigma(\boldsymbol{\ell}_{i,j})\in[0,1]^{L}$.
Discrete binary onsets $\hat{\mathbf{r}}_{i,j}\!\in\!\{0,1\}^{L}$ are obtained via the straight-through Bernoulli estimator
\begin{equation}\label{eq:st_bernoulli}
  b_{i,j,k}\sim\mathrm{Bernoulli}(q_{i,j,k}),
  \qquad
  \hat{r}_{i,j,k} = b_{i,j,k} + q_{i,j,k} - \mathrm{sg}(q_{i,j,k}),
\end{equation}
where $\mathrm{sg}(\cdot)$ denotes stop-gradient.
Because every nonzero entry of $\hat{\mathbf{r}}_{i,j}$ is binary, convolution with $\tilde{\mathbf{d}}_j$ reduces to a superposition of pure temporal shifts, so $\tilde{\mathbf{d}}_j$ appears \emph{verbatim} in every reconstruction wherever it is placed.

\paragraph{Compositional Synthesis.}
Stacking the binary onset rows into the placement matrix $\hat{\mathbf{R}}_i\in\{0,1\}^{M\times L}$, a per-timestep winner-take-all gate $\hat{\mathbf{g}}_i\in\{0,1\}^{M\times L}$ resolves residual ambiguity when several primitive types overlap in time, ensuring each timestep is owned by a single primitive identity.
The compositional synthesis is then
\begin{equation}
\label{eq:decoder}
  \hat{\mathbf{x}}_i \;\triangleq\; \mathcal{D}_{\tilde{\mathbf{D}}}(\hat{\mathbf{R}}_i)
  \;=\; \sum_{j=1}^{M}\bigl(\tilde{\mathbf{d}}_j\ast\hat{\mathbf{r}}_{i,j}\bigr)\odot\hat{\mathbf{g}}_{i,j}
  \;\in\;\mathbb{R}^{C\times L}.
\end{equation}

\begin{figure}[t]
  \centering
  \includegraphics[width=0.99\linewidth]{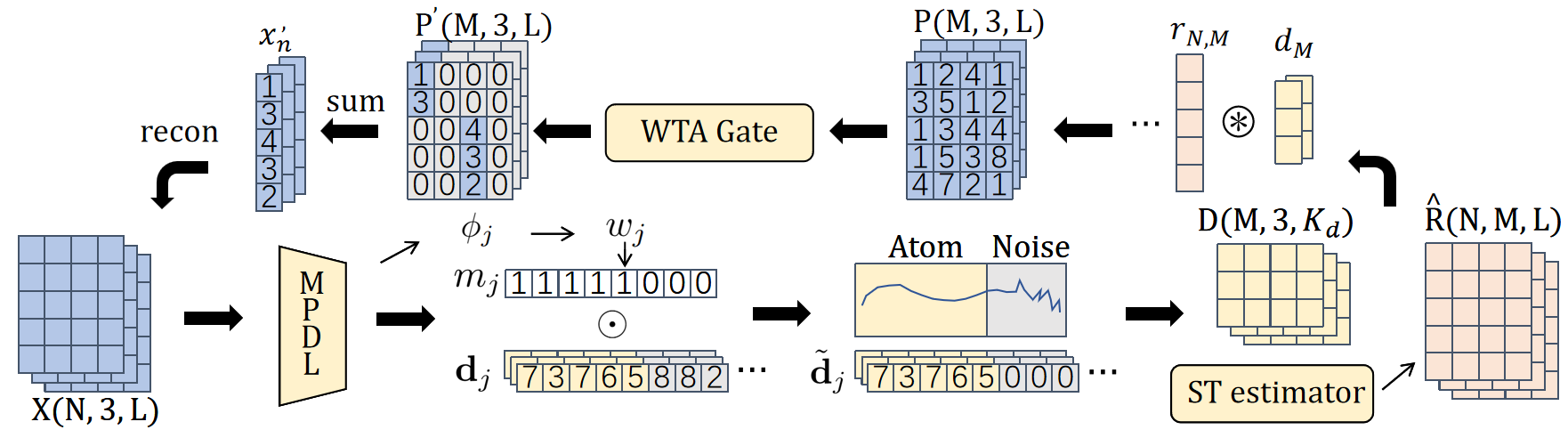}
  \caption{\textbf{Motion-primitive dictionary learning.}
    Length masks define crisp, learnable temporal boundaries for each atom, so that the masked atom itself is the self-contained motion primitive. Multi-hot binary placement encodes only the primitive's onset occurrences in $\hat{\mathbf{R}}$. A per-timestep winner-take-all gate resolves residual temporal overlaps during reconstruction, yielding a compact library of reusable, disentangled motion primitives.}
  \label{fig:fig2}
\end{figure}

We have now specified how a placement matrix $\hat{\mathbf{R}}$ synthesizes into a trajectory; what remains is to optimize $\hat{\mathbf{R}}$, the primitives $\tilde{\mathbf{D}}$, and a generative process over $\hat{\mathbf{R}}$ jointly. 
%% ====================================================================
\subsection{Structural Sparse Flow Matching as Joint Compositional Optimization}
\label{sec:joint}
\label{sec:sparse_fm}

\paragraph{$\hat{\mathbf{R}}_1$ as the single bridge.}
We treat the binary placement matrix $\hat{\mathbf{R}}_1\!\in\!\{0,1\}^{M\times L}$ as the sole bridge between primitive learning and trajectory generation: the dictionary uses $\hat{\mathbf{R}}_1$ to reconstruct observed trajectories, while the flow matching vector field uses $\hat{\mathbf{R}}_1$ as its target endpoint.
The dictionary side produces $\hat{\mathbf{R}}_1^{\mathrm{dec}}$ via the straight-through Bernoulli sampler of Sec.~\ref{sec:dict}, supervised by the requirement that $\mathcal{D}_{\tilde{\mathbf{D}}}(\hat{\mathbf{R}}_1^{\mathrm{dec}})$ explains the observed trajectory.
The flow side produces $\hat{\mathbf{R}}_1^{\mathrm{flow}} = \mathbf{Z}_t + (1-t)\,\mathbf{v}_\theta(\mathbf{Z}_t,t)$ as its predicted clean endpoint along the noise-to-placement transport.
In the single training loop introduced below, both estimates participate in the same legality energy with no stop-gradient between them: they are not independent objectives, but two parameterizations of the same combinatorial event sequence, and convergence forces them to agree.

\paragraph{Placement legality energy.}
We unify all constraints on placements through a single \emph{placement legality energy} $\Psi$, whose four sub-terms are not independent regularizers but four facets of the same event sequence induced by $\hat{\mathbf{R}}$:
\begin{equation}\label{eq:psi}
  \Psi(\hat{\mathbf{R}};\,\tilde{\mathbf{D}},\mathbf{x})
  \;=\;
  \underbrace{\Psi_{\mathrm{rec}}}_{\substack{\text{Reconstruction}\\\text{fidelity}}}
  \;+\;\lambda_s\!\underbrace{\Psi_{\mathrm{sparse}}}_{\substack{\text{event}\\\text{budget}}}
  \;+\;\lambda_p\!\underbrace{\Psi_{\mathrm{prim}}}_{\substack{\text{primitive}\\\text{integrity}}}
  \;+\;\lambda_g\!\underbrace{\Psi_{\mathrm{geo}}}_{\substack{\text{event}\\\text{geometry}}} .
\end{equation}
In plain terms, $\Psi_{\mathrm{rec}}$ asks whether the placement reconstructs the trajectory, $\Psi_{\mathrm{sparse}}$ caps the event count, $\Psi_{\mathrm{prim}}$ keeps each primitive responsible for a single coherent motion fragment, and $\Psi_{\mathrm{geo}}$ requires adjacent primitives to meet geometrically; we now give each sub-term in turn.
\textbf{Reconstruction fidelity:} $\Psi_{\mathrm{rec}}=\bigl\|\mathbf{x}-\mathcal{D}_{\tilde{\mathbf{D}}}(\hat{\mathbf{R}})\bigr\|_F^2$ requires that the placement reconstruct back to the observed trajectory through the synthesis of Eq.~\eqref{eq:decoder}.
\textbf{Event budget:} $\Psi_{\mathrm{sparse}}=\|\hat{\mathbf{R}}\|_1$ caps the number of events that may explain a trajectory.
\textbf{Primitive integrity:} writing the per-timestep gate as an ownership distribution $\boldsymbol{\pi}_{i,t}=\hat{\mathbf{g}}_{i,:,t}\in\Delta^{M-1}$ and weighting by trajectory-conditional continuity $\omega_{i,t}(\mathbf{x})=\exp(-\|\mathbf{x}_{i,t+1}-\mathbf{x}_{i,t}\|/\tau)$, $\Psi_{\mathrm{prim}}
  =\sum_{t=1}^{L-1}\omega_{i,t}(\mathbf{x})\,\bigl\|\boldsymbol{\pi}_{i,t+1}-\boldsymbol{\pi}_{i,t}\bigr\|_1.
$
This is not a generic smoothness penalty: it is conditioned on the trajectory's own kinematic continuity, so a single coherent motion fragment cannot be split across multiple atoms, exactly the meaning of ``primitive integrity''. \textbf{Event geometry.}
Each nonzero entry of $\hat{\mathbf{R}}$ defines an event $e=(j_e,k_e,P_e)$ with realized probability $P_e=\hat{R}_{j_e,k_e}$, onset $k_e$, offset $k_e+\bar{w}_{j_e}$, start point $\mathbf{a}_e=\tilde{\mathbf{d}}_{j_e}(:,0)$, and end point $\mathbf{b}_e=\tilde{\mathbf{d}}_{j_e}(:,\bar{w}_{j_e}\!-\!1)$.
The pairwise compatibility cost
\begin{equation}\label{eq:event_cost}
  c(e,e')
  \;=\;
  \underbrace{\bigl\|\mathbf{b}_e-\mathbf{a}_{e'}\bigr\|_2^2}_{\text{spatial continuity}}
  \;+\;\eta\underbrace{\bigl|k_{e'}-(k_e+\bar{w}_{j_e})\bigr|}_{\text{temporal contiguity}}
  \;+\;\rho\underbrace{\mathrm{ovl}(e,e')}_{\text{exclusivity}},
\end{equation}
expresses, in one expression, the spatial and temporal continuity that adjacent events should respect together with their mutual exclusivity. The pairwise temporal overlap $\mathrm{ovl}(e,e')$ is the length of the interval intersection of the two events on the timeline; its formal definition, properties, and differentiable surrogate are given in App.~\ref{app:ovl}. Aggregating with a vacancy-weighted sum over event pairs gives
\begin{equation}\label{eq:psi_geo}
  \Psi_{\mathrm{geo}}(\hat{\mathbf{R}};\,\tilde{\mathbf{D}})
  \;=\;
  \sum_{e<e'} P_e P_{e'}
  \underbrace{\!\!\prod_{u:\,k_e<k_u<k_{e'}}\!\!(1-P_u)}_{\text{vacancy factor}}
  \,c(e,e').
\end{equation}
The vacancy factor activates the constraint only between events that are likely to be temporally adjacent, redirecting it whenever an intermediate event intervenes. Concretely, $\prod_{u:\,k_e<k_u<k_{e'}}(1-P_u)$ is the probability that no other event occupies the open interval between $e$ and $e'$, so $c(e,e')$ contributes its full magnitude only when $e$ and $e'$ are likely to be temporal neighbors on the timeline.

\paragraph{Flow matching as energy-coupled generation.}
With placements assembled into $\hat{\mathbf{R}}_1^{\mathrm{dec}}$ at the current step, the flow interpolant and target velocity are
\begin{equation}\label{eq:interpolant}
  \mathbf{Z}_t = (1-t)\,\mathbf{Z}_0 + t\,\hat{\mathbf{R}}_1^{\mathrm{dec}},
  \quad
  \dot{\mathbf{Z}}_t = \hat{\mathbf{R}}_1^{\mathrm{dec}} - \mathbf{Z}_0,
  \quad
  \mathbf{Z}_0\!\sim\!\mathcal{N}(\mathbf{0},\sigma^2\mathbf{I}),\;\; t\!\sim\!\mathcal{U}(0,1),
\end{equation}
and the velocity field $\mathbf{v}_\theta$ regresses $\dot{\mathbf{Z}}_t$ onto $\hat{\mathbf{R}}_1^{\mathrm{dec}}$, which is not a precomputed target but the placement inferred by the dictionary side at the current step, so gradients flow through it into both $\tilde{\mathbf{D}}$ and $\boldsymbol{\ell}$ (full discussion in App.~\ref{app:full_pipeline}).
That endpoint estimate
\begin{equation}\label{eq:endpoint}
  \hat{\mathbf{R}}_1^{\mathrm{flow}} \;=\; \mathbf{Z}_t + (1-t)\,\mathbf{v}_\theta(\mathbf{Z}_t,t)
\end{equation}
enters $\Psi$ in exactly the same way as $\hat{\mathbf{R}}_1^{\mathrm{dec}}$, so geometric validity feeds back into $\theta$.
Architectural details of $\mathbf{v}_\theta$, including a duration-aware token initialization $\mathbf{h}_j^{(0)} = \mathrm{Linear}(\mathbf{Z}_{t,j,:}) + \psi(w_j) + \mathrm{MLP}(t)$ that exposes the underlying 1-D bin-packing geometry to multi-head self-attention, are deferred to App.~\ref{app:velocity_network}.

\begin{figure}[t]
  \centering
  \includegraphics[width=0.85\linewidth]{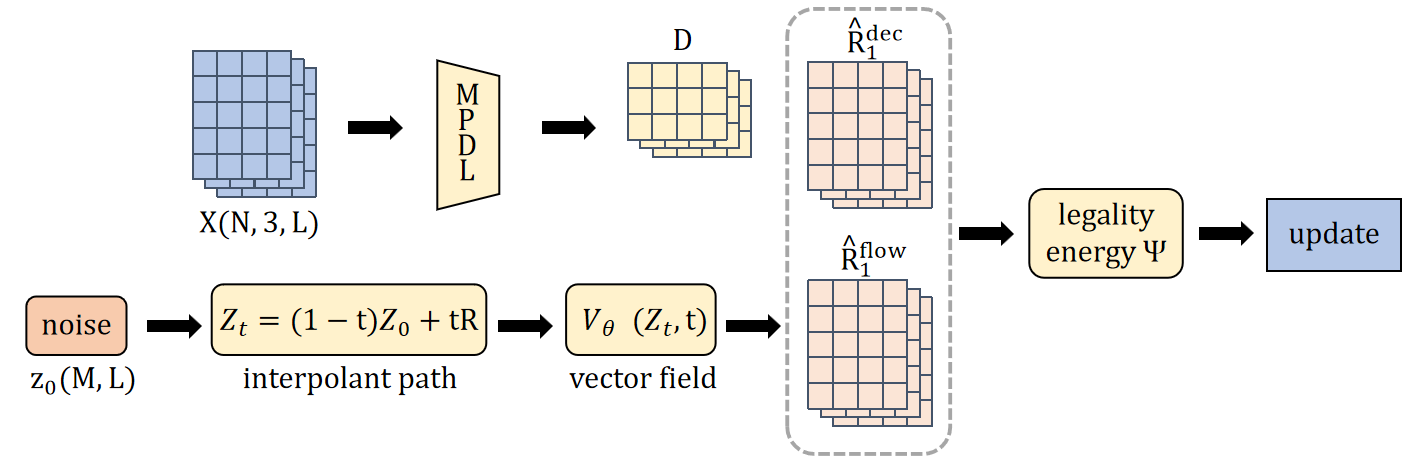}
  \caption{\textbf{Joint compositional optimization.}
    Dictionary and flow matching share a single intermediate variable $\hat{\mathbf{R}}_1$. The dictionary side produces $\hat{\mathbf{R}}_1^{\mathrm{dec}}$, which the $\mathcal{D}_{\tilde{\mathbf{D}}}$ maps back to the observed trajectory; the flow side transports noise to a predicted endpoint $\hat{\mathbf{R}}_1^{\mathrm{flow}}$. Both estimates are penalized by the same legality energy $\Psi$, and all parameters $\{\tilde{\mathbf{D}},\boldsymbol{\ell},\boldsymbol{\gamma},\theta\}$ are updated through a single computation graph.
  }
  \label{fig:fig3}
\end{figure}

\paragraph{Joint objective.}
Combining the dictionary-side legality, the flow-matching residual, and the flow-side legality, the joint training objective is
\begin{equation}\label{eq:joint}
 \min_{\tilde{\mathbf{D}},\,\boldsymbol{\ell},\,\boldsymbol{\gamma},\,\theta}
  \;\;\mathbb{E}_{\mathbf{x},\,t,\,\mathbf{Z}_0}\!\!\left[\,
    \Psi(\hat{\mathbf{R}}_1^{\mathrm{dec}};\,\tilde{\mathbf{D}},\mathbf{x})
    \;+\;\bigl\|\mathbf{v}_\theta(\mathbf{Z}_t,t)-\dot{\mathbf{Z}}_t\bigr\|_2^2
    \;+\;\beta\,\Psi(\hat{\mathbf{R}}_1^{\mathrm{flow}};\,\tilde{\mathbf{D}},\mathbf{x})
  \,\right].
\end{equation}
The three terms are not independent modules: the first shapes $\tilde{\mathbf{D}}$ and $\boldsymbol{\ell}$ so that inferred placements explain real trajectories under valid event geometry; the second is the standard flow-matching consistency, but onto a target that is itself being learned; the third closes the loop by requiring the flow's predicted endpoints to satisfy the same geometry, so improvements in the vector field feed back into the dictionary and vice versa.

%% ====================================================================
\subsection{Conditional Sampling and Prediction}
\label{sec:cond}

For trajectory prediction we condition on an observed prefix $\mathbf{x}_{1:T_{\mathrm{obs}}}$ and optional context $c$ (e.g.\ task label, environment embedding), summarized by a context vector $\mathbf{h}=f_{\mathrm{enc}}(\mathbf{x}_{1:T_{\mathrm{obs}}},c)$.
The vector field becomes $\mathbf{v}_\theta(\mathbf{Z}_t,t,\mathbf{h})$; the placement legality energy $\Psi$ and the joint objective in Eq.~\eqref{eq:joint} are otherwise unchanged.
Training remains a single loop: $\hat{\mathbf{R}}_1^{\mathrm{dec}}$ now explains the observed prefix, $\hat{\mathbf{R}}_1^{\mathrm{flow}}$ predicts the future, and $\Psi$ requires both estimates to be kinematically valid concatenations onto the same timeline.
At inference, integrating $\mathbf{v}_\theta(\cdot,\cdot,\mathbf{h})$ from $t{=}0$ to $t{=}1$ yields $\hat{\mathbf{R}}_1^{\mathrm{flow}}$, which is binarized at $0.5$ and synthesized through $\mathcal{D}_{\tilde{\mathbf{D}}}$ to produce the predicted future trajectory $\hat{\mathbf{x}}_{T_{\mathrm{obs}}+1:L}$.
Conditional prediction is thus a conditioned instance of the same joint compositional system, not a separately trained model.
Architectural details of $f_{\mathrm{enc}}$, the AdaLN conditioning of $\mathbf{v}_\theta$, classifier-free guidance, and the conditional sampling procedure are given in App.~\ref{app:conditional}.

\section{Experiments}
\label{sec:exp}
In this section, our goal is: 1) to verify that our approach achieves strong performance across diverse embodied benchmarks; 2) to analyze how different trajectory number scales and different parameter choices affect the performance of our method. Additional experimental analyses and discussions of partial failure cases are provided in detail in the appendix.
%1.消融实验
%2.M对于指标的影响
%3.字典大小对于收敛速度的影响
%4.可视化，可解释性
\subsection{Conditional Case}%ADEFDEJSD公式
\subsubsection{Embodied Trajectory Prediction in Laboratory Settings}
\label{sec:4.1lab-settings}

\paragraph{Experimental Setup}

The Open X-Embodiment\cite{open_x_embodiment_rt_x_2023} dataset is currently the largest open-source real-world robot dataset. We select a subset of manipulation tasks, including but not limited to table wiping, hanging objects, and grasping toys, in order to evaluate the applicability of our method in indoor and daily-life scenarios. We employ a non-overlapping sliding window to generate samples. Each sample contains 20 trajectory points, with the first 8 points serving as the observed trajectory and the remaining 12 points as the ground truth future trajectory. The segments for each scenario are then randomly divided into training, validation, and test sets with an 8:1:1 ratio. To probe the effect of data scale and task diversity, we construct three groups of increasing difficulty (Table~\ref{tab:task_groups}), varying the number of task categories and the total number of trajectories while keeping the sliding-window protocol fixed.

\paragraph{Baselines}

\begin{wraptable}{r}{0.48\linewidth}
\centering
\vspace{-1.0em}
\caption{Task groups used for the Open X-Embodiment experiments, spanning easy/medium/hard settings.}
\label{tab:task_groups}
\setlength{\tabcolsep}{3pt}
\footnotesize
\begin{tabular}{lcccc}
\toprule
Group & \#Tasks & \#Samples & Avg.\ len & Max len \\
\midrule
1 (easy)   & 1 & 5{,}866   & 48.33  & 64  \\
2 (medium) & 3 & 51{,}627  & 121.71 & 236 \\
3 (hard)   & 5 & 233{,}741 & 252.74 & 744 \\
\bottomrule
\end{tabular}
\vspace{-0.8em}
\end{wraptable}
We compare our approach against representative baselines from different methodological categories, selecting strong-performing methods from each category. Specifically, the baselines include CompDiffuser\cite{luo2025generative}, a diffusion-based method; MoFlow\cite{11095073}, a flow-matching-based method; 3DMoTraj\cite{10.5555/3780338.3783536}, an LSTM-based method; and LaM-SLidE\cite{sestak2026lamslide}, an encoder--decoder-based method. We additionally include ARMD\cite{Gao_Cao_Chen_2025} as a representative time-series forecasting baseline.

\begin{table}[h]
\centering
\caption{Task Group Performance Comparison. 
ADE/FDE are reported for each task group.}
\label{tab:performance_bigtable}
\setlength{\tabcolsep}{6pt}

\begin{tabular}{lcccccc}
\toprule
\multirow{2}{*}{\textbf{Method}} &
\multicolumn{2}{c}{\textbf{Task Group = 1}} &
\multicolumn{2}{c}{\textbf{Task Group = 2}} &
\multicolumn{2}{c}{\textbf{Task Group = 3}} \\
\cmidrule(lr){2-3}\cmidrule(lr){4-5}\cmidrule(lr){6-7}
& \textbf{ADE} & \textbf{FDE} & \textbf{ADE} & \textbf{FDE} & \textbf{ADE} & \textbf{FDE} \\
\midrule
ARMD\cite{Gao_Cao_Chen_2025}         & 0.952539 & 0.983366 & 0.272674 & 0.284185 & 0.278311 & 0.288450 \\
3DMoTraj\cite{10.5555/3780338.3783536}       & 0.023835 & 0.032546 & 0.020756 & 0.038789 & 0.015767 & 0.032026 \\
MoFlow\cite{11095073}       & 0.016893 & 0.031199 & 0.012383 & \underline{0.014987} & 0.015366 & 0.016302 \\
CompDiffuser\cite{luo2025generative}& 0.010486 & 0.018471 & 0.013773 & 0.029211 & 0.011172 & 0.025775 \\
LED\cite{10203059}          & 0.009302 & 0.015185 & 0.012759 & 0.023544 & 0.008919 & 0.018100 \\
LaM-SLidE\cite{sestak2026lamslide}    & \underline{0.008470} & \underline{0.013080} & \underline{0.011940} & 0.015320 & \underline{0.007841} & \underline{0.011282} \\
Ours         & \textbf{0.007424} & \textbf{0.010567} & \textbf{0.007993} & \textbf{0.011205} & \textbf{0.006333} & \textbf{0.008919} \\
\bottomrule
\end{tabular}
\end{table}

\begin{figure}[t]
  \centering
  \includegraphics[width=0.99\linewidth]{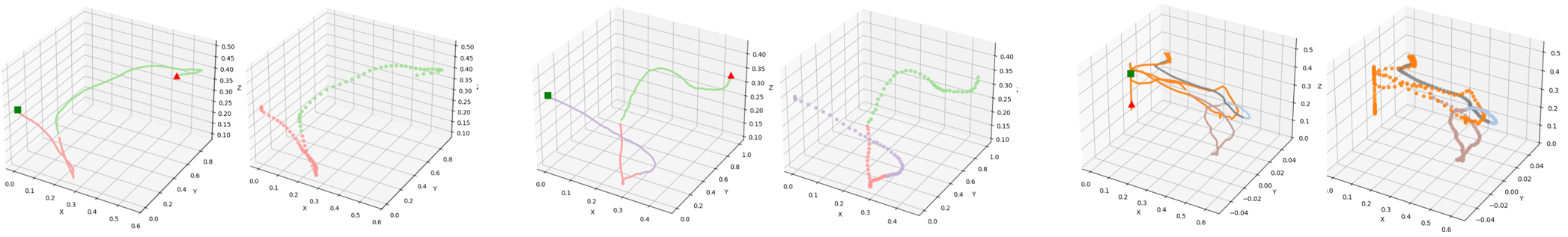}
  \caption{Here we demonstrate how primitive segments constitute trajectories. From left to right, three tasks of increasing complexity are presented. For each group, the left panel visualizes the primitive segments, while the right panel shows the resulting trajectory points.}
  \label{fig:fig4}
\end{figure}

\paragraph{Performance gain increases as task complexity increases.}
Our method achieves the best performance on all three task groups, and its relative advantage grows systematically as task difficulty increases. ARMD\cite{Gao_Cao_Chen_2025}, a generic time-series forecaster designed under stationarity assumptions, lags behind all other baselines by a wide margin, likely because embodied trajectories are non-stationary, task-conditioned compositional objects rather than smooth statistical sequences. Among the trajectory-specific baselines, compared with the strongest one, LaM-SLidE\cite{sestak2026lamslide}, our method yields a 12.3\% relative improvement in ADE and a 19.2\% relative improvement in FDE on Task Group 1. On Task Group 2, the relative improvement in ADE further expands to 33.1\%, while the FDE improvement reaches 26.9\%. On the most challenging Task Group 3, the relative improvements remain substantial at 19.2\% in ADE and 21.0\% in FDE. This trend is not incidental. As trajectory length and task diversity increase, black-box generative models are forced to fit pointwise dynamics over a much larger distributional space. By contrast, our method's complexity does not grow linearly with trajectory length, but is instead governed by the combinatorial complexity of primitive composition. 

\subsubsection{Embodied Trajectory Prediction in Natural Environments}
\label{sec:4.1.2lab-settings}
\paragraph{Dataset Introduction.} The 3DMoTraj\cite{10.5555/3780338.3783536} dataset comprises eight scenarios featuring three-dimensional (3D) trajectories collected from unmanned underwater vehicles (UUVs) operating in ocean environments. The 3D trajectories are recorded using positioning devices installed on the UUVs. Each scenario includes three UUVs following predefined formations. Notably, in addition to the complex 3D motion patterns of each UUV, the trajectories also exhibit fluctuations caused by ocean currents, which introduce significant challenges to the prediction task. We believe that the experiments in this section help validate our method's performance in embodied environments influenced by external physical conditions.

\paragraph{Experimental Setup.} On this dataset, we are committed to adopting the same experimental setup as the original authors to ensure fair comparative experiments. For this dataset, we employ the same sliding window as Sec.~\ref{sec:4.1lab-settings}. The fragments for each scenario are then randomly divided into training, validation, and test sets with a 1:1:1 ratio.

\begin{table}[h]
\centering
\caption{Scenario Performance Comparison (ADE/FDE)}
\label{tab:performance_comparison}
\setlength{\tabcolsep}{3pt}
\footnotesize% 缩小列间距，默认值通常是6pt。你可以尝试更小的值如 3pt。
% 如果4pt还不够，可以考虑将字体缩小一号：\small 或 \footnotesize
% \small % 缩小表格字体
% 或者如果以上方法都不够，可以使用 adjustbox 宏包的 resizebox 命令强制缩放表格
% \resizebox{\linewidth}{!}{%
\begin{tabular}{lccccccccc}
\toprule
\textbf{Scenario} & \textbf{\#1} & \textbf{\#2} & \textbf{\#3} & \textbf{\#4} & \textbf{\#5} & \textbf{\#6} & \textbf{\#7} & \textbf{\#8} & \textbf{Mean} \\
\midrule
SSTGCNN\cite{9156583}  & 2.86/5.19 & 1.05/1.47 & 1.08/1.65 & 2.23/3.83 & 1.15/1.34 & 2.25/4.00 & 4.49/8.28 & 2.18/3.30 & 2.16/3.63 \\
MSRL\cite{Wu_Wang_Zhou_Duan_Hua_Tang_2023}  & 3.72/5.42 & 0.62/0.73 & 0.56/0.61 & 1.69/2.05 & 1.85/2.50 & 1.54/2.84 & 1.14/1.73 & 0.90/1.15 & 1.50/2.13 \\
FlowChain\cite{10376537}  & 1.44/3.20 & 0.61/0.99 & 0.62/1.02 & 0.90/1.76 & 0.54/0.84 & 1.31/2.52 & 1.18/2.53 & 0.93/1.84 & 0.94/1.84 \\
PECNet\cite{10.1007/978-3-030-58536-5_45}  & 0.79/1.05 & 0.70/1.29 & 0.74/1.26 & 1.29/2.46 & 0.41/0.58 & 1.16/1.65 & 0.92/1.53 & 1.22/2.21 & 0.90/1.50 \\
LBEBM\cite{9578888}  & 0.72/0.98 & 0.52/0.80 & 0.64/1.11 & 0.94/1.87 & 0.35/0.57 & 1.02/1.67 & 1.27/2.35 & 1.25/2.42 & 0.84/1.47 \\
NPSN\cite{9879462}  & 0.75/0.83 & 0.69/1.21 & 0.71/0.96 & 0.83/\underline{1.28} & 0.34/\underline{0.40} & 0.97/1.20 & 0.71/\underline{0.93} & 1.03/1.50 & 0.75/1.04 \\
TrajCLIP\cite{NEURIPS2024_8cd23ec6}  & 0.56/0.94 & 0.40/0.69 & 0.37/0.70 & 1.32/2.81 & 0.33/0.58 & 0.79/1.27 & 1.04/2.02 & 0.85/1.69 & 0.71/1.34 \\
CausalHTP\cite{9710303}  & 0.69/1.29 & 0.45/0.78 & 0.45/0.78 & 0.79/1.40 & 0.49/0.87 & 1.11/1.95 & 0.97/1.96 & 0.74/1.40 & 0.71/1.30 \\
MS-TIP\cite{pmlr-v235-chib24a}  & 0.62/1.16 & 0.61/1.16 & 0.58/1.15 & \underline{0.78}/1.42 & 0.57/1.11 & 0.84/1.50 & 0.77/1.44 & 0.79/1.52 & 0.70/1.31 \\
MRGTraj\cite{10226250}  & 0.63/1.29 & \textbf{0.34}/\underline{0.54} &\underline{ 0.33}/\underline{0.55} & 0.94/1.89 & 0.42/0.83 & 0.99/1.73 & 1.06/2.44 & 0.84/1.63 & 0.69/1.36 \\
S-Implicit\cite{10.1007/978-3-031-20047-2_27}  & 0.54/0.87 & 0.43/0.72 & 0.47/0.84 & 1.13/2.31 & 0.40/0.68 & 0.85/1.38 & \underline{0.62}/1.06 & 0.98/1.92 & 0.68/1.22 \\
3DMoTraj\cite{10.5555/3780338.3783536}  & \textbf{0.36}/\textbf{0.51} & 0.37/0.60 & 0.48/0.86 & 0.81/1.69 & \textbf{0.28}/0.44 & \underline{0.69}/\underline{1.10} & 1.12/1.95 & \underline{0.55}/\underline{0.99} & \underline{0.58}/\underline{1.02} \\
\midrule
Ours & \underline{0.53}/\underline{0.55} & \underline{0.36}/\textbf{0.44} & \textbf{0.32/0.34} & \textbf{0.42/0.50} & \underline{0.32}/\textbf{0.34} & \textbf{0.50/0.51} & \textbf{0.46/0.48} & \textbf{0.36/0.37} & \textbf{0.41/0.44} \\
\bottomrule
\end{tabular}
% } % 结束 resizebox
\end{table}

\paragraph{Structured generation on 3DMoTraj.}
Beyond absolute error, the FDE/ADE ratio separates the two modeling paradigms: pointwise generators accumulate error along the temporal axis, so the endpoint deviation grows disproportionately (3DMoTraj\cite{10.5555/3780338.3783536} 1.76, S-Implicit 1.79, FlowChain 1.96), whereas our method reaches only $1.07$ ($0.44/0.41$), close to the ideal $1.0$, because it generates in the primitive--time placement space under the event-geometry term $\Psi_{\mathrm{geo}}$ (Eq.~\eqref{eq:psi_geo}) that penalizes both spatial discontinuities and temporal gaps at primitive boundaries. The 3DMoTraj setting also stresses the paradigm along an external-physics axis absent from the lab-manipulation tasks of Sec.~\ref{sec:4.1lab-settings}, namely persistent ocean-current perturbations; under these disturbances our FDE still stays below $0.55$ in every scenario (as low as $0.34$), indicating that primitive-level composition, together with the pairwise compatibility cost of Eq.~\eqref{eq:event_cost}, preserves endpoint consistency even when the agent must continually adapt to an uncontrollable force field.

% in preamble:
% \usepackage{booktabs}
% \usepackage{multirow} % 可选
% \usepackage{ulem}     % 可选（如果你想用 \uline）
% 注意：\underline 默认就能用，无需额外宏包

\subsection{Unconditional Generation}
\label{sec:uncond}
Unconditional generation provides a stricter test of whether a model captures the intrinsic motion distribution of embodied trajectories, rather than merely extrapolating from an observed prefix. In this setting, we reuse the dataset and preprocessing pipeline from Sec.~\ref{sec:4.1lab-settings}, but remove all conditioning inputs and train on full-length trajectories. For each sample, we first draw a trajectory length from the empirical length distribution of the dataset, and then require each method to generate a complete trajectory of the same length from its own generative prior. We evaluate distributional fidelity using Jensen--Shannon divergence (JSD) between the generated and real trajectory distributions under different generation scales.

\begin{figure}[t]
\centering

\begin{minipage}[t]{0.48\linewidth}
\centering
\captionof{table}{Unconditional generation performance measured by JSD. Lower is better. Generation Scale is percentage of generated data from original trajectories.}
\label{tab:jsd_unconditional}
\vspace{0.1em}

\begin{tabular}{lccc}
\toprule
& \multicolumn{3}{c}{Generation Scale} \\
\cmidrule(lr){2-4}
Method & 5\% & 10\% & 20\% \\
\midrule
ARMD\cite{Gao_Cao_Chen_2025}          & 0.653 & 0.648 & 0.647 \\
3DMoTraj\cite{10.5555/3780338.3783536}        & 0.357 & 0.324 & 0.315 \\
MoFlow\cite{11095073}        & 0.311 & 0.293 & 0.285 \\
CompDiffuser\cite{luo2025generative} & 0.186 & 0.175 & 0.172 \\
LaM-SLidE\cite{sestak2026lamslide}     & 0.173 & 0.151 & 0.148 \\
LED\cite{10203059}           & \underline{0.147} & \underline{0.130} & \underline{0.114} \\
Ours          & \textbf{0.109} & \textbf{0.096} & \textbf{0.087} \\
\bottomrule
\end{tabular}
\end{minipage}
\hfill
\begin{minipage}[t]{0.48\linewidth}
\centering

\vspace{0em} % 调这个值，让图片和表格主体对齐
\includegraphics[width=\linewidth]{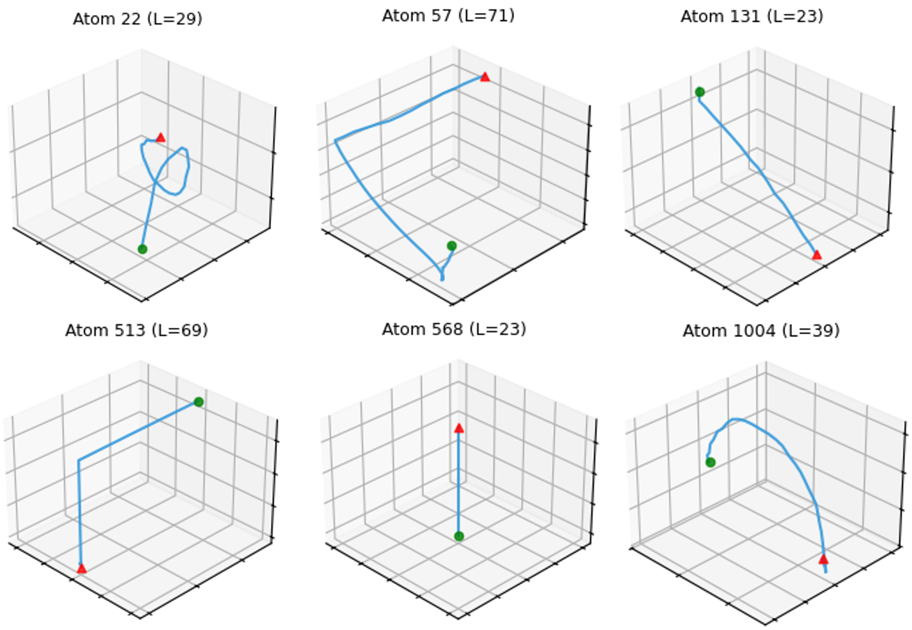}

\captionof{figure}{Randomly sampling dictionary items for primitive visualization. Clearly discernible and unambiguous semantics are observed.}
\label{fig:fig5}
\end{minipage}

\end{figure}

As shown in Table~\ref{tab:jsd_unconditional}, our method achieves the lowest JSD at all generation scales, with a clear margin over all baselines. The advantage persists as the number of generated samples increases, with JSD decreasing from $0.109$ at 5\% scale to $0.087$ at 20\% scale, indicating that the improvement is not a sampling artifact but reflects a more faithful model of the underlying trajectory distribution. We attribute this gain to the structural design of our framework: instead of generating dense coordinate sequences directly in continuous space, our method first synthesizes sparse primitive-time compositions and then maps them through a reusable motion library with explicit temporal boundaries. This compositional generation process better preserves the geometric and temporal regularities of real embodied motion, leading to substantially higher distribution-level fidelity in the unconditional setting.

\subsection{Ablation Studies}
\label{sec:ablation}

We ablate both the dictionary size $M$ and the core structural components of our method. In this setting, we reuse the dataset and preprocessing pipeline from Sec.~\ref{sec:4.1.2lab-settings} to comprehensively examine the impact of the dictionary size \(M\) and core structural components on performance across different environments. Results are reported in Table~\ref{tab:m_comparison}.

\begin{table*}[htbp]
\centering
\caption{Ablation results on dictionary size $M$ and key design components (ADE/FDE).}
\setlength{\tabcolsep}{4pt}
\resizebox{\textwidth}{!}{
\begin{tabular}{lccccccccc}
\toprule
Scenario & \#1 & \#2 & \#3 & \#4 & \#5 & \#6 & \#7 & \#8 & Mean \\
\midrule
$M=3N$            & 0.53 / 0.55 & 0.36 / 0.44 & 0.32 / 0.34 & 0.42 / 0.50 & 0.32 / 0.34 & 0.50 / 0.51 & 0.46 / 0.48 & 0.36 / 0.37 & 0.41 / 0.44 \\
$M=2N$            & 0.70 / 0.71 & 0.50 / 0.59 & 0.47 / 0.53 & 0.61 / 0.72 & 0.42 / 0.44 & 0.69 / 0.70 & 0.62 / 0.65 & 0.51 / 0.52 & 0.56 / 0.61 \\
$M=N$             & 1.07 / 1.08 & 0.69 / 0.79 & 0.66 / 0.74 & 0.86 / 1.04 & 0.60 / 0.62 & 1.11 / 1.14 & 0.92 / 0.95 & 0.75 / 0.81 & 0.83 / 0.90 \\
w/o Mask          & 0.96 / 0.97 & 0.69 / 0.80 & 0.70 / 0.81 & 0.73 / 0.86 & 0.57 / 0.60 & 0.88 / 0.90 & 0.81 / 0.83 & 0.70 / 0.74 & 0.76 / 0.81 \\
w/o Primitives    & 3.55 / 3.59 & 2.05 / 2.33 & 1.96 / 2.38 & 2.48 / 2.89 & 1.80 / 1.93 & 2.94 / 3.08 & 2.71 / 2.92 & 2.64 / 2.89 & 2.52 / 2.75 \\
\bottomrule
\end{tabular}
}
\label{tab:m_comparison}
\end{table*}

\paragraph{Dictionary size and component ablations.}
Scaling $M$ from $N$ to $3N$ monotonically improves mean ADE/FDE from $0.83/0.90$ to $0.56/0.61$ to $0.41/0.44$, as a larger search space lets pruning retain more specialized primitives for finer tiling; even at $M=N$ our FDE ($0.90$) already beats the strongest baseline in Table~\ref{tab:performance_comparison}, ruling out overparameterization and indicating that the learned atoms are reusable structural units rather than memorized fragments. The component ablations confirm where the gain comes from: removing the length mask degrades mean ADE/FDE to $0.76/0.81$, showing that explicit temporal support is necessary for sharp primitive boundaries, while removing primitive decomposition altogether collapses performance to $2.52/2.75$, so the improvement is driven by generation in the sparse compositional space rather than by the flow model alone.

\section{Discussion and Conclusion}%理论证明如收敛等写在future work

\paragraph{Limitations.}
Two practical boundaries: 1) The primitive library's quality is bounded by training-data coverage, so under-represented motion modalities yield weak or absent atoms. 2) The kinematic loss enforces only local spatial/temporal continuity, not global dynamic feasibility such as torque limits or collision avoidance, which may demand domain-specific post-processing in safety-critical deployment. Detailed analysis and failure cases are reported in the appendix.

\paragraph{Conclusion.}
We recast embodied trajectory generation as sparse composition in the primitive-time placement space: MPDL yields a compact library of self-contained motion units with explicit temporal boundaries, and Structural Sparse Flow Matching generates trajectories as combinatorial layouts over this library under differentiable event-geometry constraints.
On large-scale robot manipulation and 3D ocean benchmarks this delivers state-of-the-art accuracy, an FDE/ADE ratio of $1.07$, and gains that grow with task complexity, properties dense generative models structurally cannot achieve, suggesting embodied motion is better modeled as a composition to be assembled than a waveform to be regressed.

\bibliographystyle{unsrt} % 或 unsrt, abbrvnat 等（看你想要的格式）
\bibliography{ref}

%%%%%%%%%%%%%%%%%%%%%%%%%%%%%%%%%%%%%%%%%%%%%%%%%%%%%%%%%%%%

\end{document}